\documentclass[review]{elsarticle}

\usepackage{lineno,hyperref}
\modulolinenumbers[5]

\journal{Journal of \LaTeX\ Templates}

\usepackage{numcompress}
\begin{document}

\begin{frontmatter}

\title{Camera-Based Physiological Sensing: \\ Challenges and Future Directions}
\tnotetext[mytitlenote]{Fully documented templates are available in the elsarticle package on \href{http://www.ctan.org/tex-archive/macros/latex/contrib/elsarticle}{CTAN}.}

\author{Xin Liu, Shwetak Patel}
\address{Paul G. Allen School of Computer Science \& Engineering} \address{University of Washington, Seattle, WA, USA}
\author{Daniel McDuff}
\address{Microsoft Research, Redmond, WA, USA}

\begin{abstract}

Numerous real-world applications have been driven by the recent algorithmic advancement of artificial intelligence (AI). Healthcare is no exception and AI technologies have great potential to revolutionize the industry. Non-contact camera-based physiological sensing, including remote photoplethysmography (rPPG), is a set of imaging methods that leverages ordinary RGB cameras (e.g., webcam or smartphone camera) to capture subtle changes in electromagnetic radiation (e.g., light) reflected by the body caused by physiological processes. Because of the relative ubiquity of cameras, these methods not only have the ability to measure the signals without contact with the body but also have the opportunity to capture multimodal information (e.g., facial expressions, activities and other context) from the same sensor. However, developing accessible, equitable and useful camera-based physiological sensing systems comes with various challenges. In this article, we identify four research challenges for the field of camera-based physiological sensing and broader AI driven healthcare communities and suggest future directions to tackle these. We believe solving these challenges will help deliver accurate, equitable and generalizable AI systems for healthcare that are practical in real-world and clinical contexts. 

\end{abstract}

\begin{keyword}
\texttt{elsarticle.cls}\sep \LaTeX\sep Elsevier \sep template
\MSC[2010] 00-01\sep  99-00
\end{keyword}

\end{frontmatter}
\section{What is Non-Contact Camera-Based Physiological Sensing?}

\begin{figure*}[t!]
  \includegraphics[width=\textwidth]{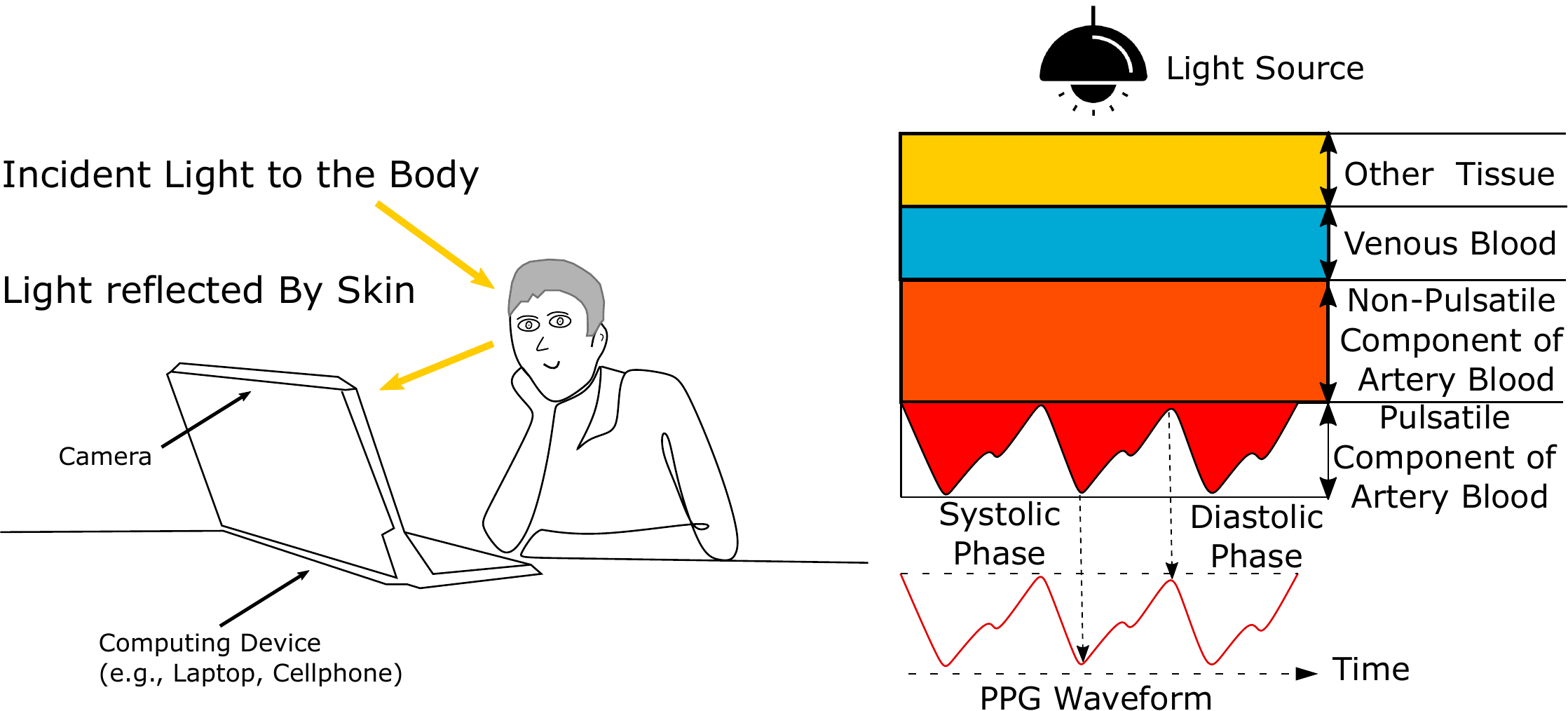}
  \caption{The principals behind camera-based physiological sensing. The volumetric changes of blood in the surface of the skin cause changes in light absorption and reflection which is the source of PPG signal. }
  \label{fig:ppg_waveform}
\end{figure*}

In the field of non-contact camera-based physiological sensing, we work on developing computational methods for extracting physiological signals (e.g., pulse rate, respiration rate, blood oxygenation, blood pressure) based on videos of the human body. In principle, as the Figure \ref{fig:ppg_waveform} is illustrated, these methods use pixel information to quantify visible light, or other electromagnetic radiation (e.g., infrared or thermal), reflected from the body. This reflected radiation is modulated by motions of the body and absorption characteristics of the skin \citep{blazek2000near,takano2007heart,Verkruysse:08,poh2010non, lam2015robust,xu2014robust, wang2016algorithmic}. In this article, we will focus primarily on the use of visible light, due to the ubiquitous nature of visible light imagers, or cameras.

As visible light penetrates between 4 to 5 mm below the skin's surface, it is modulated by the volume of oxygenated and deoxygenated hemoglobin enabling the measurement of the peripheral blood volume pulse (BVP) via photoplethysmography (PPG). The frequency channels offered by multiband (e.g., RGB) cameras enable the composition of blood, including the oxygen saturation to be measured. In addition, these pixels are affected by the motion as a person breaths in and out and by the mechanical effects of the heart beating, enabling the measurement of breathing signals and the ballistocardiogram (BCG). Analyzing the morphology of these signals, and combining them together, offers the possibility of measuring correlates of blood pressure.

In this article we will explore some of the emerging opportunities for these methods and discuss some of the challenges that need to be addressed.

\section{Applications of Non-Contact Camera-Based Physiological Sensing}

The SARS-CoV-2 (COVID-19) pandemic has accelerated the pace of change in healthcare services. In particular, how healthcare services are delivered around the world has needed to be rethought in the presence of new risks to patients and providers and restrictions on travel. The virus has been linked to the increased risk of cardiopulmonary (heart and lung related) illness with symptoms such as respiratory distress syndrome, myocarditis, and the associated chronic damage to the cardiovascular system. Experts suggest that particular attention should be given to cardiovascular protection during treatment of COVID-19 \citep{zheng2020covid}. While measurement is not the sole solution to these problems, they have acutely highlighted the need for scalable and accurate physiological monitoring.  Ubiquitous or pervasive health sensing technology could help patients conduct daily screenings, monitor the effects of medication on their symptoms, and help clinicians make more informed and accurate decisions.

The potential advantages that video-based contactless measurement offers have helped to draw a significant amount of attention to the field in recent years. Contact biomedical sensors (e.g., electrocardiograms, pulse oximeters) are the standard used for clinical screening and at-home measurement. However, these devices are usually bulky and are still not ubiquitously available, especially in low-resource settings. On the other hand, non-contact camera-based physiological sensing presents a new opportunity for highly scalable and low-cost physiological monitoring through ordinary cameras (e.g., webcams or smartphone cameras) \cite{poh2010advancements}. Besides the convenience and potential scalability, this technology could also reduce the risk of infection for vulnerable patients and discomfort caused by obtrusive leads and electrodes~\cite{villarroel2019non}. Finally, we believe there are two specifically compelling advantages of cameras over contact sensors. The first, is that they can capture multi-modal signals, including but not limited to, the activity of the subject, their appearance, facial expressions and gestures, motor control and context. One reason this helps is that the physiological measurements can be interpreted in context. For example, if someone appears in pain, an elevated heart rate can be interpreted differently than without in pain. Secondly, cameras are spatial sensors allowing for the measurement of signals from multiple parts of the body to be measured concomitantly, presenting greater opportunities for characterizing vascular parameters such as pulse transit time.

We would also argue that camera-based physiological sensing could be an influential technology in telehealth. Current telehealth procedures are mainly telephone or video-based communication services where patients see their physician or healthcare provider via Cisco Webex, Zoom or Microsoft Teams. Performing clinical visits at home increases the efficiency of clinical visits and helps people who live in remote locations. There is still a debate over whether high-quality care can be delivered over telehealth platforms. One of notable issues with current telehealth systems is that there is no way for physicians to assess patient's physiological states. The development of accurate and efficient non-contact camera-based physiological sensing technology would provide remote physicians access to the physiological data to make more informed clinical decisions.

\section{Is Non-Contact Camera-Based Physiological Sensing Applicable in Real-World Settings?}

Although there are a number of strengths to camera-based physiological sensing, we argue that there remain several technical challenges that need to be overcome before it reaches its potential. First, while there are applications in fitness tracking, most other applications require clinical-grade precision and the regulatory approval that comes with developing sensors for use in healthcare. However, there is a significant gap in accuracy between clinical-grade contact sensors and non-contact camera-based physiological sensing especially in real-world settings. Second, extending the range of vital signs that camera-based methods can measure will be important for driving adoption in many applications. To date, evaluation has focused on ``coarse'' metrics such as pulse rate and respiration rate and only preliminary work has attempted rigorous validation of pulse rate variability, left ventricle ejection time, pulse transit time and oxygen saturation and other clinically meaning metrics. 
Finally, both the hardware and software currently used for camera-based measurement has built in assumptions and data driven parameters that can and do encode biases.

\section{Existing Methods}

\begin{figure*}[t!]
  \includegraphics[width=\textwidth]{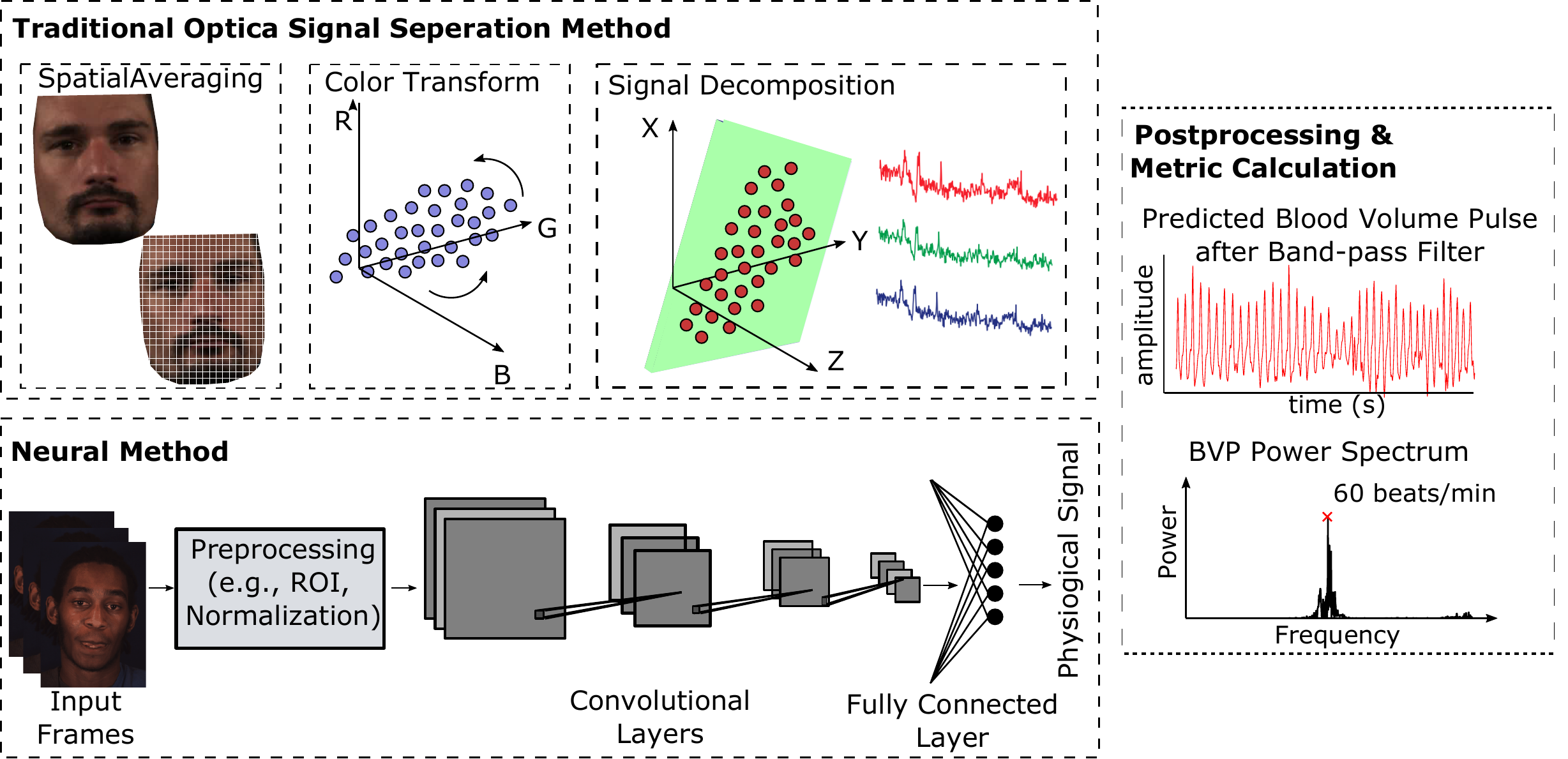}
  \caption{A general overview of traditional signal separation methods and neural network based methods}
  \label{fig:algorithm}
\end{figure*}

\subsection{Optical Principals}

Non-contact camera-based physiological sensing operates on the principle that physiological processes change how light is absorbed by, and reflected from the body. These changes are subtle but can contain rich information, including measuring the effects of blood volume changes several millimeters below the outer surface of the skin.
Most work bases computational methods on physical models. The Lambert-Beer law (LBL) \cite{lam2015robust,xu2014robust,chen2018deepphys} and Shafer's dichromatic reflection model (DRM) \cite{wang2016algorithmic} are two such models which provide a framework for capturing the effects of an imager, lighting, body motions and physiological processes on recorded pixel intensities. Given the optical characteristics of oxygenated and deoxygenated blood, we also have priors on the wavelengths of light that contain the strongest or weakest pulsatile information. This prior knowledge is important for measuring physiological parameters accurately. Most computational methods are built upon this grounding.

\subsection{Algorithms}

Many computational approaches for recovering physiological signals from videos have similar steps as the Figure \ref{fig:algorithm} shows. The first, typically involves localizing a region of interest within each video frame. In a large majority of cases the face or head are the region of interest and therefore facial detection and/or landmark detection are used. However, in other cases skin segmentation might be preferred. Aggregating pixels spatially is a subsequent step that has been used to help to reduce noise from camera quantization errors. The operation can be performed by downsampling an image~\cite{chen2018deepphys} or simply averaging all pixel values with a region of interest~\cite{poh2010non}. Many cameras capture frames from more than one frequency band (e.g., RGB) that provide complementary measurements to capture different properties of the light reflected from the body. This information can be used in two ways: 1) for understanding the composition of the blood (e.g., oxygen saturation), 2) improving the signal-to-noise ratio of the recovered blood volume pulse. Typically, computational methods leverage multiple bands and learn linear or non-linear signal decompositions to estimate the pulse waveform. This manipulation of the color channel signals can be grounded in the optical properties of the skin~\cite{wang2016algorithmic,de2013robust} or learned in a data-driven manner given a specialized learning criteria or loss function.

However, more recently supervised machine learning has become the most popular approach.  Specifically, deep learning and convolutional neural networks provide the current state-of-the-art results. These methods present the opportunity for more ``end-to-end'' learning and researchers have gradually tried to replace handcrafted processing steps with learnable components. Since the relationship between underlying physiological signal and skin pixels in a video is complicated, deep neural networks have shown superior performance on modeling such non-linear relationship compared to traditional source separation methods \citep{chen2018deepphys,yu2019remote,zhan2020analysis,lee2020meta,liu2020multi,niu2019rhythmnet,niu2020video,song2021pulsegan,lu2021dual}. Moreover, due to the flexibility of neural network, researchers have also explored neural based methods for real-time on-device inference \citep{liu2020multi}, self-supervised learning \cite{gideon2021way, lee2020meta, wang2021self}, few-shot personalization \citep{liu2021metaphys} and estimating new vital measurement such as blood pressure \cite{schrumpf2021assessment}.

\section{Validating in Clinical Settings}

To date, the large majority of non-contact camera-based physiological measurement research has involved the development and evaluation of algorithms using data of healthy subjects in research labs. There are some notable exceptions~\cite{aarts2013non}, but these are in the minority. While the research as a whole has contributed much to the understanding of the fundamental challenges associated with recovering subtle physiological signals from videos, it is also fundamentally limited. The existing datasets, especially those that are publicly available, by and large do not include irregular or problematic vitals such as atrial fibrillation and other forms of arrhythmia, low oxygen saturation levels (i.e., below 85\%) or high blood pressure. It is important that validation of the performance of non-contact camera-based system against gold-standard measurements considers the full spectrum of physiological states that could be expected, even if only rarely. 
To do this, deploying the system in clinical settings is necessary. 
The most successful examples of such deployments have been in neonatal intensive care units~\cite{aarts2013non}. We believe that another promising direction in clinical validation is deploying non-contact camera-based systems in tele-cardiology contexts. Validating non-contact solutions in clinical settings could open the door for future applications, such as large-scale screening for arrhythmia, but in order to do so there needs to be high confidence in the precision of these approaches \cite{aarts2013non, villarroel2014continuous}.
How much positive impact and improvement in telehealth visits that non-contact camera-based system can generate is an interesting question for the entire community. Validating non-contact solutions in clinical settings will open the doors for numerous future clinical applications. 

We envision more facial video data with synchronized gold-standard physiological signal will be collected in clinical settings in the future. An example could be deploying a non-contact camera-based physiological sensing system in an intensive care unit to monitor patients vitals while collecting ground-truth vital signals from the clinical-grade contact sensors. Validating non-contact solutions in clinical settings will open the doors for numerous future clinical applications.

\section{Exploring New Clinically Valuable Measures}

Non-contact camera-based physiological sensing has been studied for almost twenty years, however, most of the research still only focuses on extracting the cardiac pulse signal. Because of that, most published research usually use pulse rate measurement as the primary metric for evaluation. While modern contact wearable devices can now extract vitals like heart rate, heart rate variability, oxygen saturation, blood pressure and body temperature, non-contact camera-based sensing has been moving slowly to explore new clinically valuable measures. More recently, researchers started looking at extracting pulse transit time, pulse rate variability and oxygen saturation via an RGB camera. 

We encourage the research community to work on validation beyond average cardiac pulse measurement. For instance, preliminary research \cite{schrumpf2021assessment, slapnivcar2019blood, liu2017cuffless} 
have shown that there is a connection between blood pressure and the morphological characteristics of the photoplethysmogram (PPG). Extracting accurate systolic and diastolic blood pressure  using an RGB-camera would be an extremely valuable tool. Moreover, promising results have been shown for the measurement of oxygen saturation \cite{ding2018measuring, sun2021robust}. With the recent advancement of deep learning and computer vision techniques, we envision oxygen saturation measurement is a tractable problem for non-contact physiological sensing.

\section{Improving Equitability and Fairness}

\label{sec: fairness}

Making machine learning systems equitable is important. Non-contact camera-based physiological sensing is not an exception. Indeed, since physiological sensing involves both sensors and software, bias can be introduced in multiple parts of the design. Most consumer cameras have been designed to capture lighter skin types more effectively. Affluent, Western and Asian markets have been the primary consumers that camera companies have targeted. Thus, existing hardware present inherent biases to people with both very dark and also, although to a lesser extent, very light skin types. Furthermore, cameras have not been optimized for physiological measurement specifically. 
Melanin content impacts the intensity of light reflected from the skin and the amount of light captured by a camera. Therefore, capturing subtle pixel changes from darker skin is more challenging due to low intensity of light reflected and lower signal-noise ratio. Finally, the data distribution of existing datasets also introduces significant bias. Most of video based datasets (e.g., UBFC~\cite{bobbia2019unsupervised}) were collected from people with lighter skin types. While some of the datasets include videos of people with darker types (e.g., MMSE-HR~\cite{zhang2016multimodal}), the amount of data is still limited. 

Collecting large-scale high-quality video and physiological data from a diverse population is a potential solution to address the challenges associated with data distributions. However, recruiting and instrumenting participants is expensive, and revealing the identity and demographic information does present challenges when it comes to releasing large datasets to the research community. Therefore, we argue that developing algorithmic innovations is a more practical way to improve the fairness and equitability. Prior work have established how algorithmic innovations can help decrease skin type bias presented by computer vision systems. For example, few-shot learning can be used to personalize machine learning models and make them more robust to differences in appearance~\cite{liu2021metaphys}. With only 18 seconds of unlabeled video data, the accuracy of the neural model was significantly boosted, especially for subjects with higher melanin content in their skin. Another promising direction is data augmentation. Researchers have explored using computer graphics techniques to generate high-fidelity  avatars to synthesize synthetic video data with large diversity in appearance \cite{mcduff2020advancing, mcduff2021synthetic}. Generative neural networks have also been used to perform skin augmentation on existing datasets to generate data with diverse skin types \cite{ba2021overcoming}. We believe algorithmic innovation will help us to close the existing performance gaps in non-contact camera-based physiological sensing.

\section{Becoming Robust to Real-World Noise}

Due to the distance between camera and body skin (e.g., face), there are various sources of noise that impact the performance of non-contact camera-based sensing. Non-contact camera-based measurement  sensitive to environmental differences in the intensity and composition of the incident light. Contextual differences, including body motions in a video, also introduce large pixel changes. Under such interference, non-contact camera-based physiological sensing systems struggle to maintain high accuracy in real-world settings. Therefore, developing a generalizable method that is robust to different types of real-world noise is needed for many practical applications.

Supervised machine learning has become the most popular approach in computer vision tasks.  Specifically, deep learning and convolutional neural networks provide the current state-of-the-art results. Since the relationship between underlying physiological signal and skin pixels in a video is complicated, deep neural networks have shown superior performance on modeling such non-linear relationship compared to traditional source separation methods \cite{chen2018deepphys,yu2019remote,zhan2020analysis, lee2020meta, liu2020multi, niu2019rhythmnet, niu2020video, song2021pulsegan, lu2021dual}. As overfitting is a common issue in machine learning, most of machine learning tasks (e.g., image classification, language translation) use large pretrained models to obtain strong performance. With rich pretrained representations, researchers have opportunities to quickly advance the field as these models often perform well on downstream tasks with only modest fine-tuning. In the field of non-contact camera-based physiological sensing, such a large pretrained model is still missing. Addressing this gap could be a significant contribution to the research community. Second, the few-shot personalization and data augmentation methods described in the section \ref{sec: fairness} could also be used in improving the generalizability.

\section{Conclusion}

In this article, we have presented several challenges and opportunities for the field of non-contact camera-based physiological sensing. Given the progress over the past decade in this field and those of computer vision and machine learning, we argue that there lies significant opportunities in the near term to take these methods and apply them to impactful real-world applications.
\linenumbers

\bibliography{mybibfile}

\end{document}